\pdfoutput=1

\documentclass[11pt]{article}

\usepackage[]{acl}

\usepackage{times}
\usepackage{latexsym}
\usepackage{amsmath}
\usepackage{amssymb}
\usepackage{amsthm}
\usepackage{graphicx}

\usepackage{svg}
\usepackage{booktabs}       
\usepackage{amsfonts}       
\usepackage{nicefrac}       
\usepackage{microtype}      
\usepackage{xcolor}         
\usepackage{multirow}
\usepackage{float}
\usepackage{amssymb}
\usepackage{amsmath}
\usepackage{bm}
\usepackage{latexsym}
\usepackage{subfigure}
\usepackage{graphicx}
\usepackage[ruled,vlined]{algorithm2e}
\usepackage{color, soul, colortbl}
\usepackage{xcolor}
\definecolor{myred}{rgb}{0.839,0.341,0.271}
\definecolor{myblue}{rgb}{0.843,0.898,0.941}
\definecolor{mygrey}{rgb}{0.976,0.976,0.976}

\definecolor{mygreen}{HTML}{FCF0D9}
\definecolor{mybackblue}{HTML}{DDF0F5}

\definecolor{themegreen}{HTML}{C5E0B4}
\definecolor{themeblue}{HTML}{CFE2F3}
\definecolor{themeyellow}{HTML}{FFE699}
\definecolor{themedarkyellow}{HTML}{FFBF87}

\newcommand\figcaption{\def\@captype{figure}\caption}
\newcommand\tabcaption{\def\@captype{table}\caption}

\usepackage{caption}

\usepackage{algpseudocode}

\usepackage{tikz}
\usepackage{makecell}
\usetikzlibrary{positioning}
\usepackage{cleveref}
\usepackage{ragged2e}
\usepackage{soul}
\usepackage{tcolorbox}
\usepackage{wrapfig}
\crefname{section}{§}{§§}
\Crefname{section}{§}{§§}

\usepackage[T1]{fontenc}

\usepackage[utf8]{inputenc}

\usepackage{microtype}

\usepackage{inconsolata}

\definecolor{mygray}{rgb}{0.976,0.976,0.976}
\usepackage[english]{babel}
\newtheorem{lemma}{Lemma}

\title{Scalable Efficient Training of Large Language Models with Low-dimensional Projected Attention}

\author{Xingtai Lv$^{1}$, Ning Ding$^{1*}$, {Kaiyan Zhang}$^{1}$, {Ermo Hua}$^{1}$, {Ganqu Cui}$^{2,3}$, {Bowen Zhou}$^{1,2}$\thanks{corresponding authors} \\
$^{1}$Department of Electronic Engineering, Tsinghua University, $^{2}$Shanghai AI Laboratory \\
$^{3}$Department of Computer Science and Technology, Tsinghua University \\
\texttt{lvxt24@mails.tsinghua.edu.cn}, \texttt{\{dn97, zhoubowen\}@tsinghua.edu.cn} \\ 
}

\begin{document}
\maketitle
\begin{abstract}

Improving the effectiveness and efficiency of large language models (LLMs) simultaneously is a critical yet challenging research goal.
In this paper, we find that low-rank pre-training, normally considered as efficient methods that will compromise performance, can be scalably effective when reduced parameters are precisely targeted.
Specifically, applying the low-dimensional module only to the attention layer -- resolves this issue and enhances both effectiveness and efficiency.
We refer to this structure as \textsl{Low-dimensional Projected Attention (LPA)} and provide an explanatory analysis.
Through extensive experimentation at parameter scales of 130M, 370M, and scaling up to 3B, we have validated the effectiveness and scalability of LPA. Our results show that LPA model can save up to 12.4\% in time while achieving an approximate 5\% improvement in test perplexity (ppl) and on downstream tasks compared with the vanilla Transformer.

\end{abstract}

\section{Introduction}

Improving large language models' (LLMs)~\cite{bommasani2021opportunities, han2021pre, brown2020language,touvron2023llama,zhou2024generative} effectiveness and efficiency simultaneously presents challenges due to inherent trade-offs, which remains a critical research goal in the research field.
Among series methods proposed to alleviate this issue, parameter-efficient fine-tuning~\cite{houlsby2019parameter,li2021prefix,zaken2021bitfit,ding2023parameter} offer valuable insights. Notably, low-rank or low-dimension techniques such as LoRA~\cite{hu2021lora} demonstrate on-par or even enhanced performance over traditional full-parameter fine-tuning with reduced computational resources.

Intuitively, besides the fine-tuning phase, adapting LoRA’s principles to the \textit{pre-training phase} through low-rank decomposition is both viable and promising, which can yield substantial benefits if effectiveness is maintained. 
However, existing studies have found that the direct low-rank pre-training often compromises the effectiveness.
To reduce such effects, strategies such as iteratively accumulating low-rank updates~\cite{lialin2023relora} or integrating low-rank decomposition directly into the gradient~\cite{Zhao2024GaLoreML} have been suggested. 
Whether it's the original LoRA or these improved methods, they all involve performing low-rank decomposition and updates on "amounts of change" (weights or gradients), and do not reduce the number of parameters in the model itself, which face obstacles in maintaining efficiency during subsequent inference and fine-tuning stages.
Therefore, an ideal scenario would be permanently reducing the number of parameters (computational load) through efficient methods, without compromising or even enhancing the performance of pre-trained models.


To achieve this goal, is it feasible to directly perform low-rank decomposition on the matrices in the model itself, rather than on the changes? 
Current limited research suggests that existing low-rank pre-training methods experience performance losses and uncertainties~\cite{lialin2023relora,Zhao2024GaLoreML}, with even fewer studies exploring more direct approaches.
However, in this paper, we demonstrate that such direct low-rank pre-training is feasible, provided that the parameters to be reduced are more \textit{precisely targeted}.
Specifically, we describe the reduction of parameters as replacing the original matrices with low-dimensional modules. We find that using low-dimensional modules in the feed-forward neural (FFN) layers or across all layers negatively impacts the model's effectiveness.
However, we observe that employing them in the attention layers consistently allows the model to outperform the original Transformer. We refer to this structure as \textsl{Low-dimensional Projected Attention (LPA)}, provide an explanation, and experimentally demonstrate its ability to reliably enhance both the efficiency and effectiveness of the model.


We validate the effectiveness of the LPA model on two Transformer model configurations, assessing both pre-training and downstream task performance. With a particular focus on the scalability of LPA model, we observe that it remains effective even when the model parameters scale up to 3B.
Furthermore, our study explores the effects of the hyperparameter on LPA, the necessity of integrating the low-dimensional module into every sublayer of the attention layer, and how to distribute any extra parameters effectively.
The code of this work will be publicly available at \url{https://github.com/TsinghuaC3I/LPA}.


\section{Related Work}

\paragraph{Low-rank Parameter-efficient Fine-tuning.} 
Parameter-efficient fine-tuning optimize only a tiny portion of parameters while keeping the majority of the neural network frozen~\cite{houlsby2019parameter,li2021prefix,lester2021power,hu2021lora,zaken2021bitfit,ding2023sparse}, saving significant time and computational costs and achieving performance comparable to full parameter fine-tuning on many tasks~\cite{ding2023parameter}.
Low-rank adaptation (LoRA) is one of the most effective and influential parameter-efficient fine-tuning methods, having found widespread application~\cite{dettmers2023qlora}.
The LoRA method involves freezing the weights $\mathbf{W}_0$ of the pre-trained model while training two low-rank decomposition matrices $\mathbf{W}_u$ and $\mathbf{W}_d$, resulting in the output of the LoRA module being represented as $\mathbf{z} \leftarrow \mathbf{W}_0\mathbf{x} + \mathbf{W}_u\mathbf{W}_d\mathbf{x}$.
We drew inspiration from LoRA and its improvement works, adapting them to the pre-training process to enhance effectiveness and efficiency of the model.

\paragraph{Low-rank Pre-training for Neural Network.}
Some efforts have focused on making pre-training more efficient by reducing the number of trainable parameters~\cite{lin2020hotcake,yuan2020growing}, and after finding that modules with low-dimension often yield poor results~\cite{bhojanapalli2020low}, many works have concentrated on combining two low-rank matrices to reduce the parameter count while keeping the module dimensionality constant~\cite{schotthofer2022low,idelbayev2020low,zhao2023inrank,thangarasa2023spdf}.
Current research has predominantly emphasized refining pre-training methods for CNN networks~\cite{sui2024elrt,jaderberg2014speeding} or employing smaller language models~\cite{kamalakara2022exploring}.
However, some studies have found that low-rank pre-training can negatively impact model performance and training effectiveness, leading to the use of low-rank updates to train high-rank networks or the introduction of low-rank decomposition in gradient for optimization~\cite{lialin2023relora,Zhao2024GaLoreML}.
Additionally, \citealt{liu2024deepseek} introduces low-rank latent states in the attention layer, successfully optimizing the KV cache.

We discover that the unsatisfactory performance of the direct low-rank pre-training stems from the lack of precise parameter reduction placement. This insight guides our further exploration into the impact of low-dimensional modules and their applications at various locations within the model on both effectiveness and efficiency.

\section{Low-dimensional Projected Attention}

We use a low-dimensional module for replacing the original weight matrix, and observe the varying effects of incorporating the low-dimensional structure in different modules.
We provide an explanatory analysis of these findings and propose the Low-dimensional Projected Attention (LPA). Additionally, we examine the efficiency of this approach.

\subsection{Low-dimensional Module}

\begin{figure*}[!ht]
    \centering
    \includegraphics[width=0.9\linewidth]{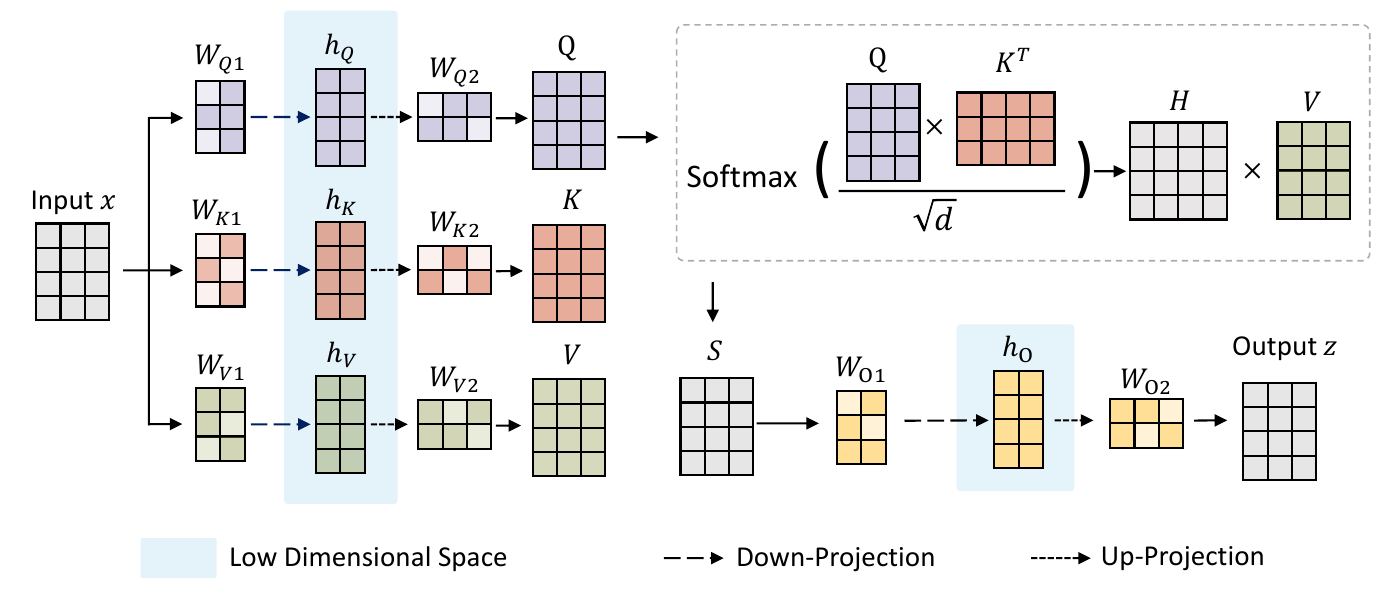 }
    \caption{An illustration of the Low-dimensional Projected Attention (LPA). The calculations in $\texttt{softmax}$ function measure the relationships between input tokens. 
    }
    \label{fig:line_plot}
\end{figure*}

The low-dimensional module is constructed by sequentially connecting two low-dimensional matrices.
Specifically, given a predetermined hyperparameter $r$, which is typically less than $\frac{d_\text{in}\times d_\text{out}}{d_\text{in}+d_\text{out}}$, the low-dimensional module comprises two matrices $\mathbf{W}_A \in \mathbb{R}^{d_\text{in} \times r}$ and $\mathbf{W}_B \in \mathbb{R}^{r \times d_\text{out}}$, where $d_\text{in}$ and $d_\text{out}$ represent the input and output dimensions of the parameter matrix, respectively.
The input data $\mathbf{x}\in \mathbb{R}^{L \times d_\text{in}}$ passes through $\mathbf{W}_A$ and $\mathbf{W}_B$ sequentially, and the forward propagation of the low-dimensional module is expressed as $\mathbf{z}\leftarrow \mathbf{W}_B\left(\mathbf{W}_A(\mathbf{x})\right)$.
The low-dimensional module is employed to displace the weight metric $\mathbf{W} \in \mathbb{R}^{d_\text{in} \times d_\text{out}}$ in linear layers of the original model, such as the weight in the Query sublayer of the attention layer.

For the classic Transformer architecture, the forward propagation formula for the original attention layer is:
\begin{equation}
\mathbf{z} \leftarrow \texttt{S} \left ( \frac{\mathbf{x}\mathbf{W}_Q\mathbf{W}_K^T\mathbf{x}^T}{\sqrt{d}} \right )\mathbf{x}\mathbf{W}_V\mathbf{W}_O,
\label{equ: original attn}
\end{equation}
where $\mathbf{W}_Q$, $\mathbf{W}_K$, $\mathbf{W}_V$ and $\mathbf{W}_O$ are the parameter matrices of the Query, Key, Value, and Output layers, $\texttt{S}$ is the $\texttt{softmax}$ function, and $d$ is the dimension of the attention layer.
When applying low-dimensional module to the attention layer, the corresponding parameters for the Query, Key, Value, and Output layers are $\mathbf{W}_{Q1}$, $\mathbf{W}_{Q2}$, $\mathbf{W}_{K1}$, $\mathbf{W}_{K2}$, $\mathbf{W}_{V1}$, $\mathbf{W}_{V2}$, $\mathbf{W}_{O1}$ and $\mathbf{W}_{O2}$, where the matrices with subscript $1$ correspond to the $\mathbf{W}_A$ matrix of the low-dimensional module, and the matrices with subscript $2$ correspond to the $\mathbf{W}_B$ matrix.
The forward propagation formula for the attention layer with the low-dimensional module is:
\begin{align}
    \mathbf{z} \leftarrow \texttt{S} \left ( \frac{\mathbf{x}\mathbf{W}_{Q1}\mathbf{W}_{Q2}\mathbf{W}_{K2}^T\mathbf{W}_{K1}^T\mathbf{x}^T}{\sqrt{d}} \right )
    \notag \\
    \mathbf{x}\mathbf{W}_{V1}\mathbf{W}_{V2}\mathbf{W}_{O1}\mathbf{W}_{O2}.
\label{equ: low attn}
\end{align}
Similarly, the forward propagation formula for the original FFN layer is:
\begin{equation}
  \mathbf{z} \leftarrow \delta \left ( \mathbf{x}\mathbf{W}_{U}\right )\mathbf{W}_{D},
  \label{equ: original ffn}
\end{equation}
where $\mathbf{W}_{U}$ and $\mathbf{W}_{D}$ are the up-projection and down-projection matrices of the FFN layer, and $\delta$ is the non-linear activation function.
When applying the low-dimensional module to the FFN layer, the corresponding parameters for the up-projection and down-projection matrices are $\mathbf{W}_{U1}$, $\mathbf{W}_{U2}$, $\mathbf{W}_{D1}$ and $\mathbf{W}_{D2}$.
The forward propagation formula for the FFN layer with the low-dimensional module is:
\begin{equation}
  \mathbf{z} \leftarrow \delta \left ( \mathbf{x}\mathbf{W}_{U1}\mathbf{W}_{U2}\right )\mathbf{W}_{D1}\mathbf{W}_{D2}.
  \label{equ: low ffn}
\end{equation}

\subsection{Position Optimization of Low-dimensional Module}
\label{Position Optimization of Low-dimension Module}

The model performance may be influenced by the position of the low-dimensional module within the model, a phenomenon akin to what has been widely observed in the field of parameter-efficient finetuning~\cite{zaken2021bitfit,hu2022sparse,zhang2023adaptive,ding2023sparse}. In order to validate this influence and ascertain the appropriate position, we apply the low-dimensional module separately in the attention layers, FFN layers, and across all layers.
The resulting models are based on the 135M and 369M Transformers, and we adjust the hyperparameter $r$ to ensure that the parameter count of these models remains approximately consistent across these three position settings.

To confirm the robustness of the optimal low-dimensional module position, we apply it in two different Transformer model settings, each containing only decoders. The \textit{Model Setting 1} employs the Layer Normalization~\cite{ba2016layer} and the "ATTN(FFN)-Norm-Add" regularization process, with ReLU~\cite{Fukushima1975CognitronAS} as the activation function. The corresponding models are pre-trained on the WikiText-103 dataset~\cite{Merity2016PointerSM}, which contains 0.1B tokens. The \textit{Model Setting 2} uses RMS Normalization and the same FFN layer as in LLaMA~\cite{touvron2023llama}, along with the "Norm-ATTN(FFN)-Add" regularization process. The corresponding models are pre-trained on the Pile dataset~\cite{gao2020pile}, using 2.6B tokens for the 130M parameter model and 6.8B tokens for the 370M parameter model.

\begin{figure}[!ht]
    \centering
    \includegraphics[width=1\linewidth]{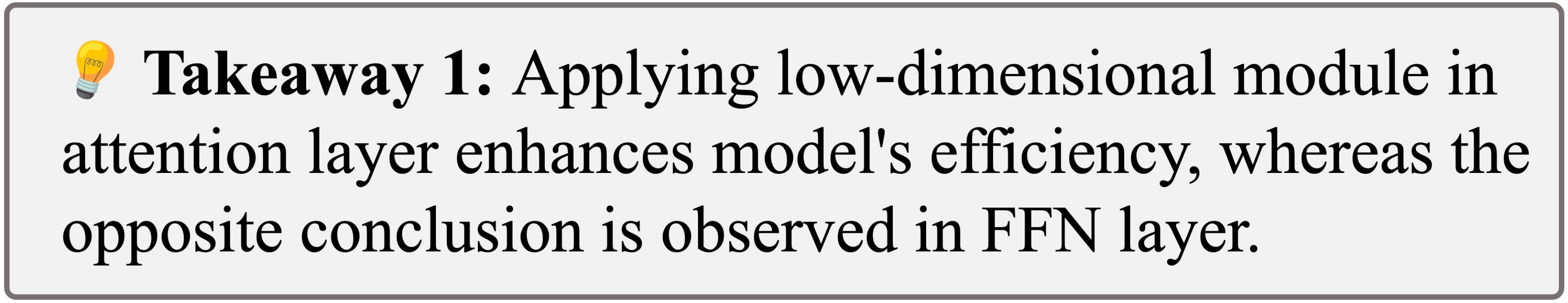}
\end{figure}

The perplexities of these pre-trained models on test datasets are presented in Table~\ref{tab:low dif module}. 
The models with low-dimensional modules employed across all layers perform worse than the original Transformers, consistent with the findings of~\citealt{lialin2023relora}. Applying the low-dimensional module to the attention layers yields a considerable improvement in pre-training performance compared to its application to FFN layers and across all layers.
Notably, for the 370M parameter model, the performance of the model with low-dimensional modules in attention layers even surpasses that of the original Transformer model, which suggests that employing the low-dimensional module in the attention layers can serve as a beneficial strategy.

\begin{table}[!htbp]
    \centering
    \scalebox{0.75}{
    \begin{tabular}{c|ccc}
    \toprule
        \textbf{Transformer} &\textbf{Low Attn} &\textbf{Low FFN} &\textbf{Low All}\\
    \midrule
        \rowcolor{mygray} \multicolumn{4}{c}{\textit{Model Setting 1}} \\ 
    \midrule
        14.61(135M) &\textbf{14.66(125M)} &15.25(125M) &15.00(126M) \\
        13.65(369M) &\textbf{12.89(319M)} &14.12(325M) &13.14(318M)\\
    \midrule
        \rowcolor{mygray} \multicolumn{4}{c}{\textit{Model Setting 2}} \\ 
    \midrule
        18.84(134M) &\textbf{18.95(115M)} &20.43(116M) &20.64(117M) \\
        12.10(368M) &\textbf{11.68(318M)} &12.77(318M) &12.68(314M)\\
    \bottomrule
    \end{tabular}}
    \caption{Test perplexities for models with low-dimensional module integration at various positions and the original Transformer models. \textbf{Low Attn}, \textbf{Low FFN}, and \textbf{Low All} separately mean applying the low-dimensional module in the attention layers, FFN layers, and across all layers. The model size is provided in parentheses.}
    \label{tab:low dif module}
\end{table}

\subsection{Explanation for Position Optimization}
\label{Low-dimension Projection Attention Transformer (LoPAT)}
Our preliminary experiments indicate that the optimal position for low-dimensional modules in the Transformer architecture is the attention layer. Further detailed observations reveal that applying low-dimensional modules to the FFN layers diminishes the model's effectiveness compared to the original Transformer model, whereas applying them to the attention layers enhances the model's performance, particularly in the 370M parameter setting.

\begin{figure}[!ht]
    \centering
    \includegraphics[width=1\linewidth]{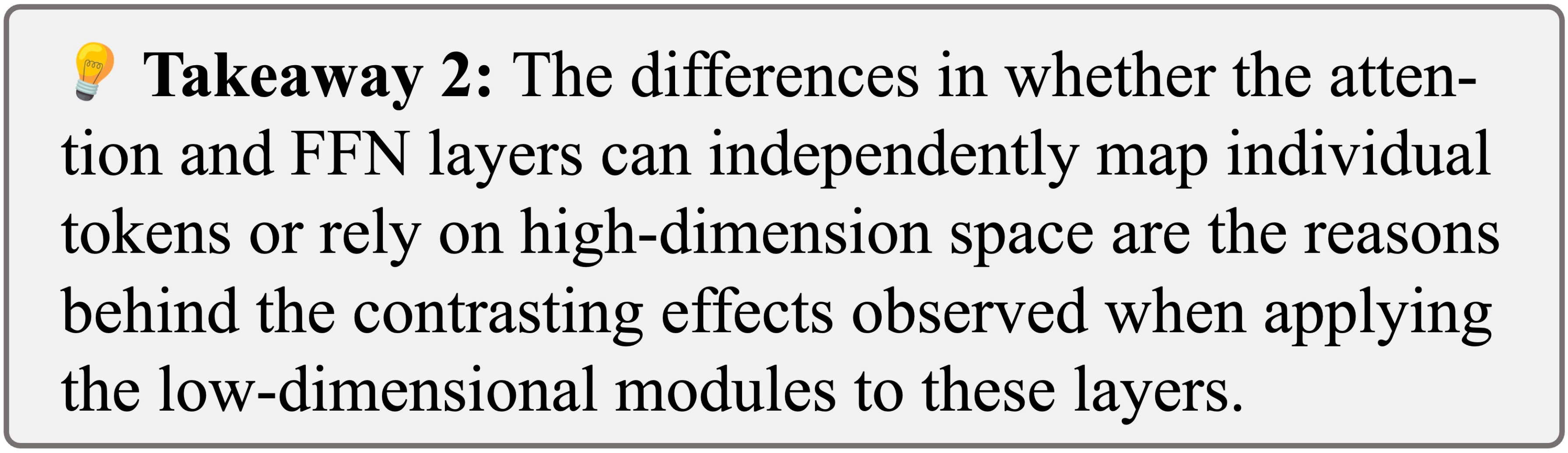}
\end{figure}

On one hand, according to Lemma~\ref{lemma: attn} and Lemma~\ref{lemma: ffn}, the attention layer cannot independently map individual tokens, whereas the FFN layer performs computations for each input token independently. On the other hand, the FFN layer typically projects inputs to a high-dimensional space, while the attention layer does not engage in similar operations. We posit these differences are the primary reasons for the positive effect of applying low-dimensional modules within the attention layer, contrasted with their negative impact in the FFN layer. Detailed empirical explanations are provided in Appendix~\ref{Explanatory Analyses for Phenomena}, based on the perspective of viewing the introduction of low-dimensional modules as a two-step projection.

\begin{lemma}
In the attention layer, for the input vector $\mathbf{x}_i \in \mathbb{R}^{1 \times d_\text{in}}$ of the $i$-th input token, the corresponding output $\mathbf{z}_i \in \mathbb{R}^{1 \times d_\text{out}}$ satisfies 
\begin{equation}
  \mathbf{z}_i \leftarrow \texttt{S} \left ( \frac{\mathbf{x}_i\mathbf{W}_Q\mathbf{W}_K^T\mathbf{x}^T}{\sqrt{d}} \right )\mathbf{x}\mathbf{W}_V\mathbf{W}_O,
\end{equation}
indicating that $\mathbf{z}_i$ is dependent on all the vectors in the input $\mathbf{x}$, especially for the computation in the Key, Value layers.
\label{lemma: attn}
\end{lemma}

\begin{lemma}
In the FFN layer, the output $\mathbf{z}_i \in \mathbb{R}^{1 \times d_\text{out}}$ corresponding to $\mathbf{x}_i \in \mathbb{R}^{1 \times d_\text{in}}$ satisfies 
\begin{equation}
  \mathbf{z}_i \leftarrow \delta \left ( \mathbf{x}_i\mathbf{W}_{U}\right )\mathbf{W}_{D},
\end{equation}
implying that $\mathbf{z}_i$ is only dependent on $\mathbf{x}_i$ instead of other vectors in the input $\mathbf{x}$.
\label{lemma: ffn}
\end{lemma}

However, when the original model has a low parameter count, applying low-dimensional modules to the attention layer degrades the effect of projection, leading to a noticeable decline in the model's capacity to fit the data.
As a result, this method is effective only for models with a larger parameter count, with a critical threshold between 130M and 370M parameters, as identified in our pre-experiments in Section~\ref{Position Optimization of Low-dimension Module}

Therefore, applying low-dimensional modules to the attention layer is the optimal strategy in Transformer models. This essentially involves two-step projection through a low-dimensional space within the attention layer, and we term this model architecture \textsl{Low-dimensional Projected Attention (LPA)}.

\subsection{Methodological Efficiency}
The core architecture of the LPA model is composed of low-dimensional modules.
Because of the lower parameter number in these modules, pre-training LPA model reduces memory consumption and is more conducive to large-scale training. 
Moreover, unlike other low-rank pre-training approaches~\cite{schotthofer2022low,lialin2023relora} and methods that involve pre-training with full parameters followed by finding the approximate low-dimensional matrices during inference~\cite{chen2021dsee}, our LPA model maintains a low-dimensional structure in both the pre-training and subsequent inference and fine-tuning stages, implying sustained efficiency throughout the entire lifecycle of the model. 
Theoretically, compared to the original linear layer, where the input $\mathbf{x} \in \mathbb{R}^{L \times d_\text{in}}$ undergoes forward computation with floating point operations (flops) at $\mathcal{O}(L \cdot d_\text{in} \cdot d_\text{out})$, utilizing the low-dimensional module reduces this to $\mathcal{O}(L \cdot r \cdot (d_\text{in}+d_\text{out}))$, considering $r < \frac{d_\text{in}\cdot d_\text{out}}{d_\text{in}+d_\text{out}}$.

\begin{figure}[!ht]
    \centering
    \includegraphics[width=1\linewidth]{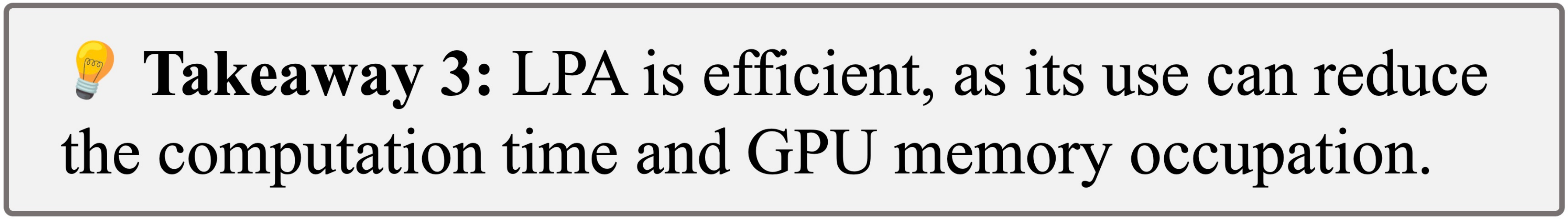}
\end{figure}

In order to experimentally verify the methodological efficiency, we conduct tests on 135M, 369M, and 3.23B Transformers with \textit{Model Setting 1} and the corresponding LPA models during the evaluation stage, measuring the clock time and GPU memory consumption on the WikiText-103 dataset (for 135M and 369M models) and the Pile dataset~\cite{gao2020pile} (for 3.23B models) with identical compute infrastructure and batch size.
Theoretically, applying low-dimensional module to the attention layers reduces flops from $8L\cdot{d_\text{in}}\cdot{d_\text{out}} + 2L^2 \cdot {d_\text{out}}$ to $8L\cdot r\cdot(d_\text{in}+d_\text{out}) + 2L^2 \cdot {d_\text{out}}$.
As presented in Table~\ref{tab:Methodological Efficiency}, both the evaluation time and GPU memory consumption of the LPA model are smaller compared to the corresponding Transformer, demonstrating the methodological efficiency. Furthermore, the LPA model offers the potential to reduce the KV cache, as the hidden states projected into the low-dimensional space can be stored in place of the KV cache.

\begin{table}[!htbp]
    \centering
    \scalebox{0.9}{
    \begin{tabular}{c|c|cc}
    \toprule
        &\textbf{Params} &\makecell[c]{\textbf{Time} \\ \textbf{pre Step}}
        &\makecell[c]{\textbf{GPU} \\ \textbf{memory}}\\
    \midrule
    Transformer &135M &153.4ms &2302MiB \\
    LPA &125M &150.6ms &2276MiB \\
    \midrule
    Transformer &369M &351.0ms &4648MiB \\
    LPA &319M &322.9ms &4464MiB \\
    \midrule
    Transformer &3.23B &6.923s &71.94GiB \\
    LPA &2.43B &6.066s &70.26GiB \\  
    \bottomrule
    \end{tabular}}
    \caption{The average evaluation time pre step and GPU memory consumption pre device for Transformer and LPA with various model sizes.}
    \label{tab:Methodological Efficiency}
\end{table}

\section{Experiments}

Extensive experiments are conducted to validate the effectiveness of LPA across models of various scales, particularly emphasizing its efficacy with the 3.23B models. 
Furthermore, we investigate the impact of hyperparameter $r$ on LPA, whether applying the low-dimensional module to all sublayers in the attention layer is necessary, and the allocation of surplus parameters.

\subsection{Effectiveness of LPA}
\label{Effectiveness of LPA}

\begin{figure*}[!ht]
    \centering
    \includegraphics[width=0.9\linewidth]{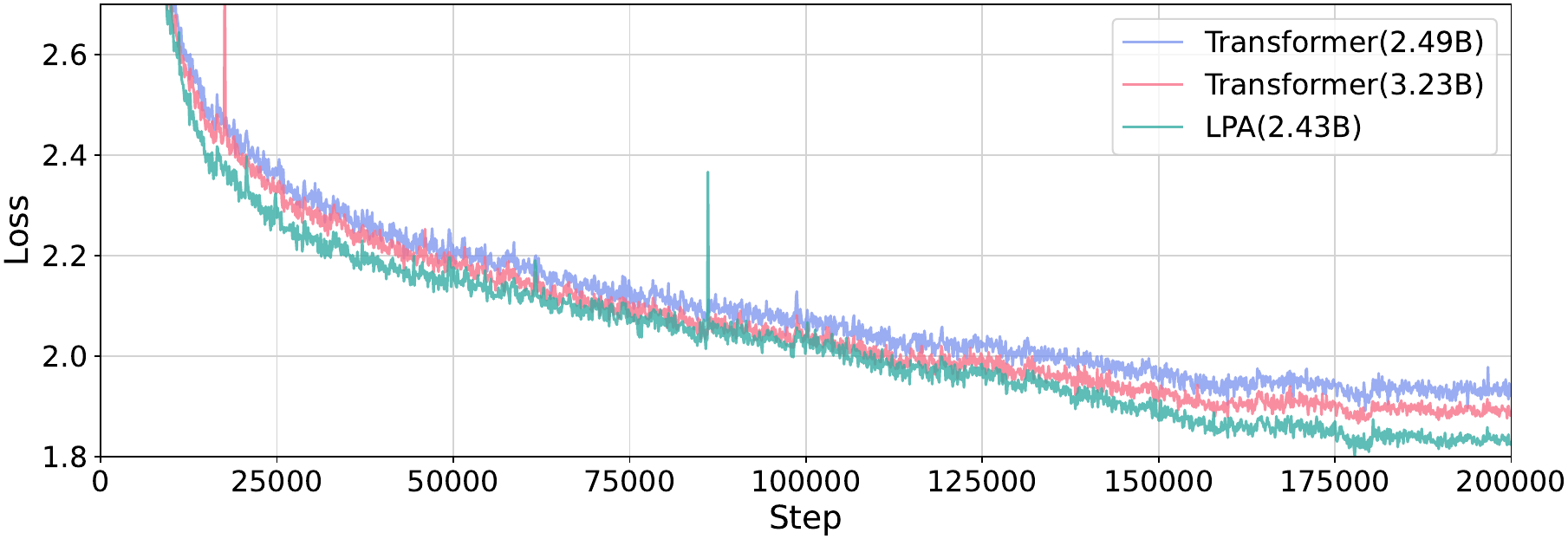}
    \caption{
    Training loss for the 2.43B LPA model, the 3.23B Same-Dim Transformer, and the 2.49B Transformer with nearly the same parameter count as the LPA model.
    }
    \label{fig:line_plot}
\end{figure*}

\paragraph{Experimental Settings.} To validate the effectiveness and robustness of the LPA architecture, we conduct experiments with two model settings introduced in Section~\ref{Position Optimization of Low-dimension Module}, pre-training models with parameter sizes of 130M and 370M.
For \textit{Model Setting 1}, we use the WikiText-103 dataset~\cite{Merity2016PointerSM}, consisting of 0.1B tokens, and set $r$ of LPA to 256. For \textit{Model Setting 2}, we pre-train the models using 2.6B tokens from the Pile dataset~\cite{gao2020pile} for the 130M parameter model and 6.8B tokens for the 370M parameter model, with the LPA architecture $r$ set to 128 or 256.
Detailed model configurations and training hyperparameters are provided in Table~\ref{tab: hyperparameters of small model} in Appendix~\ref{Hyperparameters of Model Architecture and Pre-training}. For the implementation of our models, we leverage the Huggingface Transformers~\cite{wolf2020transformers} and PyTorch~\cite{paszke2019pytorch} frameworks. Our computational infrastructure is powered by the NVIDIA GeForce RTX 3090 (maximum GPU memory=24GB), NVIDIA A800 (maximum GPU memory=80GB), and NVIDIA A6000 (maximum GPU memory=48GB).


As indicated in Table~\ref{tab: hyperparameters of small model}, the parameter count of the LPA model typically ranges from 75\% to 90\% of the corresponding Transformer, referred to as \textit{the Same-Dim Transformer}.
To compare the performance of the LPA and Transformer models under the same parameter settings, we also pre-train Transformer models with parameter counts nearly equal to those of LPA models.
For each model, repeated pre-training with 3 random seeds is performed, and following pre-training, we evaluate the models on test datasets, using perplexity (ppl) as the performance metric.


\paragraph{Results and analysis.} The mean test perplexity and standard deviation for each model are presented in Table~\ref{tab: small model ppl}. Generally speaking, the LPA model can achieve similar or slightly better performance compared to the Same-Dim Transformer. Moreover, the performance of the LPA model is notably superior to that of the Transformer with a nearly equivalent model size.
However, for the 130M parameter size model, the test perplexity of the LPA model is slightly higher than that of the Same-Dim Transformer across two model settings.
This could be attributed to the fact that with fewer parameters in the model, each parameter has to accommodate more, thus making the parameter count more crucial. The integration of low-dimensional modules into the attention layer considerably reduces the model's fitting capability, thereby diminishing overall performance. Consequently, employing LPA with 130M parameters may not enhance the model's effectiveness and may even have adverse effects.

\begin{table}[!htbp]
        \centering
        \scalebox{0.77}{
        \begin{tabular}{cc|c}
        \toprule
            \makecell[c]{\textbf{Transformer}\\\textbf{(Same-Dim)}} &\makecell[c]{\textbf{Transformer}\\\textbf{(Same-Param)}} &\textbf{LPA} \\
        \midrule
            \rowcolor{mygray} \multicolumn{3}{c}{\textit{Model Setting 1}} \\ 
        \midrule
            \textbf{14.61$_{\pm 0.16}$(135M)} &14.69$_{\pm 0.05}$(128M) &14.66$_{\pm 0.14}$(125M) \\ 
            13.65$_{\pm 0.09}$(369M) &13.75$_{\pm 0.02}$(319M) &\textbf{12.89$_{\pm 0.11}$(319M)} \\
        \midrule
            \rowcolor{mygray} \multicolumn{3}{c}{\textit{Model Setting 2}} \\ 
        \midrule
            \textbf{18.44$_{\pm 0.40}$(134M)} &19.59$_{\pm 0.12}$(116M) &19.08$_{\pm 0.12}$(115M) \\
            12.10$_{\pm 0.01}$(368M) &12.33$_{\pm 0.01}$(318M) &\textbf{11.70$_{\pm 0.02}$(318M)} \\
        \bottomrule
        \end{tabular}}
        \caption{Test perplexities for all models with parameter sizes of 130M and 370M. The model size is provided in parentheses.}
        \label{tab: small model ppl}
    \end{table}

\subsection{Scaling up to 3.23B}
\label{Scaling up to 3.23B}


In this section, experiments are conducted on the 3B-scale models, including the pre-training of a 2.43B LPA model, a 3.23B Same-Dim Transformer, and a 2.49B Transformer with nearly the same parameter count as the LPA model. Inspired by LLaMA~\cite{touvron2023llama}, we adopt the pre-normalization for these large models. Compared to pre-training smaller models, we utilize a larger dataset, specifically 13\% of the Pile dataset, amounting to 51B tokens, without data repetition during pre-training. Additional hyperparameters for the model architecture and training settings are detailed in Table~\ref{tab: hyperparameters of large model} in Appendix~\ref{Hyperparameters of Model Architecture and Pre-training}.

\begin{table}[!htbp]
    \centering
    \scalebox{0.9}{
    \begin{tabular}{cc|c}
    \toprule
        \makecell[c]{\textbf{Transformer}\\\textbf{(Same-Dim)}} &\makecell[c]{\textbf{Transformer}\\\textbf{(Same-Param)}} &\textbf{LPA} \\
    \midrule
        6.45(3.23B) &6.69(2.49B) &\textbf{6.11(2.43B)} \\
    \bottomrule
    \end{tabular}}
    \caption{Test perplexities for all models with parameter sizes of 3B. The model size is provided in parentheses.}
    \label{tab:3B model ppl}
\end{table}


Figure~\ref{fig:line_plot} illustrates the training loss for three models, and Table~\ref{tab:3B model ppl} presents their perplexities on the test set.
The 2.43B LPA model achieves a lower test perplexity than both the 3.23B and 2.49B Transformer models. Moreover, the training loss of the 2.43B LPA model consistently remains below those of the two Transformer models, particularly in the later stages of pre-training.
This indicates that the LPA maintains a significant advantage when the model parameter is scaled up to 3B, suggesting substantial potential for application in even larger models and demonstrating its scalability.


\subsection{Downstream Tasks Performance}
\label{Downstream Tasks Performance}

\begin{table*}[!ht]
        \centering
        \scalebox{0.83}{
        \begin{tabular}{c|c|ccccccccc}
        \toprule
            \textbf{Model} & \textbf{Params} & \makecell[c]{\textbf{CoLA}\\Mcc} & \makecell[c]{\textbf{SST-2}\\Acc} & \makecell[c]{\textbf{MRPC}\\Acc} & \makecell[c]{\textbf{QQP}\\Acc/F1} & \makecell[c]{\textbf{STS-B}\\Corr} & \makecell[c]{\textbf{MNLI}\\Acc(m/mm)} & \makecell[c]{\textbf{QNLI}\\Acc} & \makecell[c]{\textbf{RTE}\\Acc} & \textbf{Avg.}  \\ 
        \midrule
            Transformer  &369M  &\makecell[c]{18.28\\(0.47)} &\makecell[c]{84.94\\(0.54)} &\makecell[c]{74.35\\(3.40)} &\makecell[c]{86.60/81.95\\(0.04)/(0.08)} &\makecell[c]{72.47\\(1.14)} &\makecell[c]{71.69/71.81\\(0.37)/(0.21)} &\makecell[c]{80.92\\(0.08)} &\makecell[c]{52.76\\(0.34)} &67.47 \\
            LPA &319M &\makecell[c]{\textbf{25.46}\\\textbf{(0.66)}} &\makecell[c]{\textbf{86.51}\\\textbf{(0.99)}} &\makecell[c]{\textbf{78.92}\\\textbf{(0.69)}} &\makecell[c]{\textbf{87.44/83.06}\\\textbf{(0.11)/(0.20)}} &\makecell[c]{\textbf{78.77}\\\textbf{(0.23)}} &\makecell[c]{\textbf{73.73/74.20}\\\textbf{(0.08)/(0.47)}} &\makecell[c]{\textbf{83.26}\\\textbf{(0.45)}} &\makecell[c]{\textbf{53.60}\\\textbf{(0.36)}} &\textbf{70.72} \\
        \bottomrule
        \end{tabular}}
        \caption{Test results of the pre-trained LPA and Transformer models on the GLUE benchmark. "Mcc", "Acc", "F1" and "Corr" represent matthews correlation coefficient, accuracy, the F1 score, and pearson correlation coefficient respectively. And "Acc(m/mm)" represents the results corresponding to matched and mismatched datasets of MNLI. The standard deviation is provided in parentheses.}
        \label{tab:glue}
\end{table*}


To further demonstrate the superiority of the LPA model over the Transformer, in addition to comparing test perplexities, we also evaluate the performance of the pre-trained 369M Transformer and the 319M LPA model with \textit{Model Setting 1} on downstream tasks.
Using the GLUE benchmark~\cite{wangglue}, which is widely recognized for the natural language understanding, we conduct full-parameter fine-tuning on CoLA~\cite{warstadt2019neural}, SST-2~\cite{socher2013recursive}, MRPC~\cite{dolan2005automatically}, QQP~\cite{wangglue}, STS-B~\cite{wangglue}, MNLI~\cite{williams2017broad}, QNLI~\cite{rajpurkar2016squad} and RTE~\cite{dagan2005pascal,haim2006second,giampiccolo2007third,bentivogli2009fifth}.
We perform repeated experiments with 3 random seeds and report the average results and standard deviations in Table~\ref{tab:glue}.

Due to our use of WikiText-103 as the training dataset in \textit{Model Setting 1} and the inherent limitations of decoder-only models in classification tasks, the overall scores on the GLUE benchmark are relatively lower. WikiText-103 is a research-oriented dataset with a relatively small amount of training data and is not specifically designed for the capabilities required by the GLUE benchmark.
However, our results indicate that the pre-trained LPA model outperforms the Transformer, particularly on tasks such as MRPC and STS-B. Additionally, the standard deviation for the LPA model is not significantly different from that of the Transformer, suggesting that the observed performance improvements on the GLUE tasks can indeed be attributed to the LPA model.

\subsection{Apply LPA with Different $r$}

For the LPA, $r$ is the most critical hyperparameter, and it is essential to investigate the impact of different $r$ on the performance of the LPA models.
We pre-train a 369M Transformer with \textit{Model Setting 1} and the corresponding LPA models with $r$ set to 256, 128, 64, and 32, followed by conducting repeated experiments with 3 random seeds and computing the average test perplexity for each configuration.

\begin{figure}[!ht]
    \centering
    \includegraphics[width=1\linewidth]{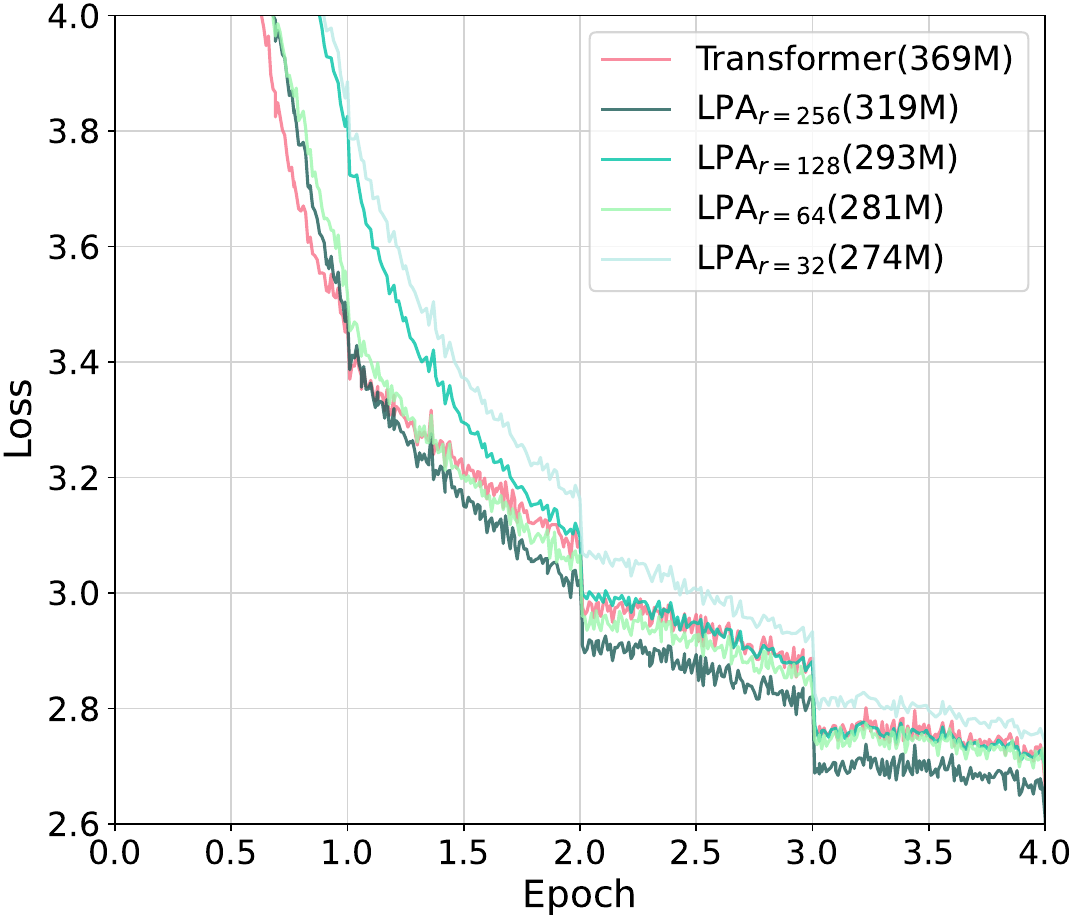}
    \caption{Training loss for Transformer and LPA models with different $r$. The darker curves correspond to larger values of $r$ in LPA.}
    \label{fig:different ranks}
\end{figure}

\begin{table}[!htbp]
    \centering
    \scalebox{0.9}{
    \begin{tabular}{c|cc}
    \toprule
        &\textbf{Param} &\textbf{Perplexity} \\
    \midrule
        \textbf{Transformer} &369M &13.65 \\
    \midrule
        \textbf{LPA$_{r=256}$} &319M &12.89 \\
        \textbf{LPA$_{r=128}$} &293M &13.03 \\
        \textbf{LPA$_{r=64}$} &281M &13.19 \\
        \textbf{LPA$_{r=32}$} &274M &13.82 \\
    \bottomrule
    \end{tabular}}
    \caption{Parameter count and test perplexities for Transformer and LPA models with different $r$.}
    \label{tab:different ranks}
\end{table}

Figure~\ref{fig:different ranks} shows the training loss curves of these models, and Table~\ref{tab:different ranks} presents the test perplexity results. Overall, although the performance of the LPA model degrades as $r$ decreases, the LPA models generally outperform the Same-Dim Transformer in both training loss and test perplexity, which indicates that the LPA model is quite tolerant to variations in $r$.
However, when the $r$ is too low, such as 32, the effectiveness of the LPA is relatively inferior compared to the Transformer, which may be because a meager $r$ results in a lack of crucial parameters, significantly impacting the model's fitting capability.


\subsection{Apply Low-dimensional Module to Different Sublayers in Attention}

In the aforementioned experiments, we apply the low-dimensional module to all sublayers of the attention layer, including the Query, Key, Value, and Output layers. In this section, we explore whether applying the low-dimensional module to only some sublayers can achieve better results.
We design combinations of sublayers to which the low-dimensional module is applied based on the functional characteristics of them. Specifically, according to Lemma~\ref{lemma: attn}, the computations in the Key and Value layers require all the vectors in the input $x$. Additionally, the Query, Key, and Value layers collectively handle the computation of the relationships between the input tokens.
Therefore, we consider two configurations in the experiments: applying the low-dimensional module to the Key and Value layers, and applying it to the Query, Key, and Value layers, which are denoted as LPA$_{K,V}$ and LPA$_{Q,K,V}$, respectively.

\begin{table}[!htbp]
    \centering
    \scalebox{0.9}{
    \begin{tabular}{c|cc}
    \toprule
            &\makecell[c]{\textit{Model}\\\textit{Setting 1}} &\makecell[c]{\textit{Model}\\\textit{Setting 2}} \\
    \midrule
        \textbf{Transformer} &13.65(369M) &12.10(368M) \\
        \textbf{LPA} &12.89(319M) &11.68(318M) \\
    \midrule
        \textbf{LPA$_{K,V}$} &13.29(344M) &11.73(343M) \\
        \textbf{LPA$_{Q,K,V}$} &12.94(331M) &11.80(330M) \\
    \bottomrule
    \end{tabular}}
    \caption{Test perplexities for LPA, LPA$_{K,V}$, LPA$_{Q,K,V}$, and the Same-Dim Transformer with parameter sizes of 370M. The model size is provided in parentheses.}
    \label{tab: different sublayers}
\end{table}

The LPA, LPA$_{K,V}$, LPA$_{Q,K,V}$, and the Same-Dim Transformer with parameter sizes of 370M and two model settings are pre-trained, and Table~\ref{tab: different sublayers} reports their test perplexities. 
We observe that the performance of both LPA$_{K,V}$ and LPA$_{Q,K,V}$ is slightly inferior to that of LPA, indicating that applying the low-dimensional module to all sublayers in the attention layer is more appropriate.

\subsection{Allocating Surplus Parameters across Modules}

The reduced parameter of the LPA model compared to the Same-Dim Transformer presents an opportunity to allocate the saved parameters to other modules of the model, which is a worthwhile avenue to explore for further enhancing the model's effectiveness.
Building upon the LPA model, we respectively allocate the parameters in three ways: 
(1) \textbf{Attn Dim.} Increasing the output dimensions of $\mathbf{W}_{Q}$, $\mathbf{W}_{K}$, $\mathbf{W}_{V}$ and the input dimensions of $\mathbf{W}_{O}$ in attention layers.
(2) \textbf{FFN Dim.} Expanding the output dimensions of the up-project matrix $\mathbf{W}_{U}$ and the input dimensions of the down-project matrix $\mathbf{W}_{D}$ in the FFN layers.
(3) \textbf{Layer Num.} Enlarging the number of layers in LPA model.
We conduct repeated experiments with \textit{Model Setting 1}, using the same training settings and 3 random seeds for the Transformer and LPA model, and the average test perplexities are presented in Table~\ref{tab:Allocating Surplus Parameters}.

\begin{table}[!htbp]
    \centering
    \scalebox{0.85}{
    \begin{tabular}{c|cc}
    \toprule
            &\makecell[c]{\textbf{130M}\\\textbf{Param Size}} &\makecell[c]{\textbf{370M}\\\textbf{Param Size}}\\
    \midrule
        \textbf{Transformer} &14.61(135M) &13.65(369M) \\
        \textbf{LPA} &14.66(125M) &12.89(319M) \\
    \midrule
        \textbf{Attn Dim.} &\textbf{14.32(135M)} &\textbf{12.85(369M)} \\
        \textbf{FFN Dim.} &14.38(135M) &13.02(369M) \\
        \textbf{Layer Num.} &14.39(138M) &13.04(371M) \\
    \bottomrule
    \end{tabular}}
    \caption{Test perplexities for variant models obtained through parameter reallocation and baselines. The model size is provided in parentheses.}
    \label{tab:Allocating Surplus Parameters}
\end{table}

Both the LPA model and the models obtained through parameter reallocation exhibit lower test perplexity compared to the Transformer, which indicates that these parameter reallocation strategies have a positive impact compared to the original Transformer model. 
Notably, the models employing the \textbf{Attn Dim.} strategy demonstrate the most favorable performance in terms of test perplexity, indicating that allocating surplus parameters to increase the dimensionality of attention layers leads to superior results, making it the most effective parameter reallocation scheme.
Furthermore, compared to LPA model, the \textbf{FFN Dim.} and \textbf{Layer Num.} models exhibit higher test perplexity at the 370M parameter size, suggesting that augmenting the FFN dimension and the layer number on top of LPA architecture may be unsuitable solutions, especially in the context of large parameter size.

\section{Conclusion}
\label{sec:Conclusion}

This paper demonstrates that low-rank pre-training can enhance both the effectiveness and efficiency of LLMs when reduced parameters are precisely targeted.
By incorporating low-dimensional modules specifically in the attention layers, we develop the Low-dimensional Projected Attention (LPA), which outperforms Transformers without the efficiency compromises. 
Our empirical analysis and experiments show that LPA maintains its effectiveness even as model parameters scale up to 3B.
Additionally, we explore the impact of hyperparameters and the optimal reallocation of surplus parameters, providing a robust framework for future enhancements in LLM pre-training.

\section*{Limitations}

Despite the encouraging results demonstrated by this paper, certain limitations in our current study are worth acknowledging.
First of all, our explanation in Section~\ref{Low-dimension Projection Attention Transformer (LoPAT)} is empirical rather than a rigorous theoretical explanation with mathematical derivation.
Furthermore, due to computational resource limitations, we conduct experiments with a 3B parameter scale on only one Transformer model setting and don't verify the effectiveness of LPA at larger parameter scales.
Last, we find that the efficiency of LPA during the pre-training phase is not very apparent, which may require the introduction of KV cache because LPA has the potential to reduce KV cache, but we don't explore this further.

\section*{Acknowledgements}
This work is supported by the National Science and Technology Major Project (2023ZD0121403), Young Elite Scientists Sponsorship Program by CAST (2023QNRC001), National Natural Science Foundation of China (No. 62406165).

\bibliography{anthology,custom}

\clearpage
\appendix

\section{Hyperparameters of Model Architecture and Pre-training}
\label{Hyperparameters of Model Architecture and Pre-training}

In this section, we present the key hyperparameters from the aforementioned experiments.
The hyperparameters for pre-training the Transformer and LPA models with parameter sizes of 130M and 370M, as described in Section~\ref{Effectiveness of LPA}, are shown in Table~\ref{tab: hyperparameters of small model}, and the hyperparameters for pre-training models with parameter sizes of 3B, as described in Section~\ref{Scaling up to 3.23B}, are listed in Table~\ref{tab: hyperparameters of large model}.
The upper and lower parts of these tables respectively display the hyperparameters related to the model architecture and pre-training settings.

\begin{table}[!htbp]
    \centering
    \scalebox{0.8}{
    \begin{tabular}{c|cc|cc}
    \toprule
        &\multicolumn{2}{c|}{\textit{Model Setting 1}} & \multicolumn{2}{c}{\textit{Model Setting 2}} \\
    \midrule
        \textbf{Params(Trans)} &135M &369M &134M &368M \\
        \textbf{Params(LPA)} &125M &319M &115M &318M \\
        \textbf{$r$} &256 &256 &128 &256 \\
        \textbf{Hidden Size} &768 &1024 &768 &1024 \\
        \textbf{Heads} &8 &8 &12 &16 \\
        \textbf{FFN Dim} &3072 &4096 &2048 &2736 \\
        \textbf{Layers} &12 &24 &12 &24 \\
    \midrule
        \textbf{lr(Trans)} &8e-4 &8e-4 &1e-3 &1e-3 \\
        \textbf{lr(LPA)} &8e-4 &8e-4 &1e-3 &8e-4 \\
        \textbf{Epoch} &10 &8 &1 &1 \\
        \textbf{Batch Size} &82K &98K &82K &61K \\
        \textbf{Seq.len.} &512 &1024 &256 &512 \\
    \bottomrule
  \end{tabular}}
    \caption{Hyperparameters of the model architecture and pre-training settings. \textbf{lr(Trans)} and \textbf{lr(LPA)} mean the learning rates for pre-training Transformer and LPA models.}
    \label{tab: hyperparameters of small model}
\end{table}

\begin{table}[!htbp]
    \centering
    \scalebox{0.8}{
    \begin{tabular}{c|ccc}
    \toprule
        &\makecell[c]{\textbf{Transformer}\\\textbf{(Same-Dim)}} &\makecell[c]{\textbf{Transformer}\\\textbf{(Same-Param)}} &\textbf{LPA} \\
    \midrule
        \textbf{Params} &3.23B &2.49B &2.43B \\
        \textbf{$r$} &- &- &512 \\
        \textbf{Hidden Size} &4096 &4096 &4096 \\
        \textbf{Heads} &32 &32 &32 \\
        \textbf{FFN Dim} &14436 &14436 &14436 \\
        \textbf{Layers} &16 &12 &16 \\
    \midrule
        \textbf{lr} &3e-4 &3e-4 &6e-4 \\
        \textbf{Epoch} &1 &1 &1 \\
        \textbf{Batch Size} &262K &262K &262K \\
        \textbf{Seq.len.} &4096 &4096 &4096 \\
    \bottomrule
  \end{tabular}}
    \caption{Hyperparameters of the model architecture and pre-training settings for large models. \textbf{lr} means the learning rate for training.}
    \label{tab: hyperparameters of large model}
\end{table}

\section{Explanatory Analyses for Phenomena Described in Section~\ref{Position Optimization of Low-dimension Module}}
\label{Explanatory Analyses for Phenomena}

There are two primary empirical explanations for the different effects when applying the low-dimensional modules to the attention layer and FFN layer.
First, the parameter matrix with low-dimensional modules can be viewed as a two-step projection, which involves first mapping the input data into a low-dimensional space and then back into the target space. Typically, the FFN layer projects the input into a high-dimensional space via $\mathbf{W}_{U}$, processes it with the non-linear activation function, and then maps it back to the original space via $\mathbf{W}_{D}$. 
The heavy reliance on the high-dimensional space of the FFN layers means that introducing low-dimensional space through low-dimensional modules negatively impacts it. 
Additionally, for each token in the input consisting of $L$ tokens, considering Lemma~\ref{lemma: attn} and $\texttt{S} \left ( \frac{\mathbf{x}_i\mathbf{W}_Q\mathbf{W}_K^T\mathbf{x}^T}{\sqrt{d}} \right ) \in \mathbb{R}^{1 \times L}$, the \texttt{softmax} computation in the attention layer results in one-dimensional weight data for $L$ tokens, indicating that the attention layer is less sensitive to the dimensionality of the input space. Hence, introducing a low-dimensional space has a minimal negative impact on the attention layer.

Secondly, for the input data which comprises $L$ tokens, based on Lemma~\ref{lemma: ffn}, the projection of these $L$ tokens in the FFN layer is independent, effectively processing them sequentially. 
In contrast, based on Lemma~\ref{lemma: attn}, the computation in the attention layer involves the relationships between each input token and all $L$ tokens.
Theoretically, since the projection can be optimized to any possible choice, projecting data into a low-dimensional space before mapping it back to the target space should not affect the size of the output space.
However, in practice, this operation tends to concentrate the output in several subspaces within the target space, reducing the output space size, which constrains the possible output values and makes it harder to identify the optimal weight point.

This negative impact is substantial for the FFN layer, but for the attention layer, the reduced output space implies that the data points for input tokens are closer together, making their relationships easier to capture. Consequently, applying the low-dimensional module to the attention layers can enhance the model's effectiveness.

\end{document}